\documentclass[10pt,journal]{IEEEtran}
\ifCLASSOPTIONcompsoc
  \usepackage[nocompress]{cite}
\else
  \usepackage{cite}
\fi
\usepackage{times}
\usepackage{threeparttable}
\usepackage{epsfig}
\usepackage{graphicx}
\usepackage{amsmath}
\usepackage{amssymb}
\usepackage{times}
\usepackage{epsfig}
\usepackage{graphicx}
\usepackage{amsmath}
\usepackage{amssymb}
\usepackage{multirow}
\usepackage{url}
\usepackage{hyperref}
\usepackage{caption}
\usepackage{float} %设置图片浮动位置的宏包
\usepackage{algpseudocode,algorithm}
\usepackage{gensymb}
\usepackage{booktabs}
\usepackage{xcolor}
\usepackage{bm} 

\usepackage{caption}
\usepackage{graphicx}
\usepackage{float}
\usepackage{algpseudocode,algorithm}
\usepackage{gensymb}
\usepackage{booktabs}
\usepackage{multirow}
\usepackage{stfloats}
\graphicspath{{figure/}}
\ifCLASSINFOpdf
\else
\fi
\ifCLASSOPTIONcompsoc
 \usepackage[caption=false,font=footnotesize,labelfont=sf,textfont=sf]{subfig}
\else
  \usepackage[caption=false,font=footnotesize]{subfig}
\fi
\begin{document}
\title{CaCo: Both Positive and Negative Samples are Directly Learnable via Cooperative-adversarial Contrastive Learning}

\author{Xiao Wang,~\IEEEmembership{Student~Member,~IEEE},
        Yuhang Huang, Dan Zeng 
        and Guo-Jun Qi,~\IEEEmembership{Fellow,~IEEE}% <-this % stops a space
\IEEEcompsocitemizethanks{\IEEEcompsocthanksitem X. Wang was with Department of Computer Science, Purdue University,West Lafayette, 47906, USA.
\IEEEcompsocthanksitem Y. Huang and D. Zeng were with the School of Communication and Information Engineering, Shanghai University, Shanghai, 200444, China. 
\IEEEcompsocthanksitem G-J. Qi was with the Futurewei Seattle Cloud Lab, Seattle, WA, 98006, USA. Email: guojunq@gmail.com.
}% <-this % stops an unwanted space
\thanks{Manuscript received March, 2022; revised xx,xx; accepted xx,xx.(Corrresponding author: Guo-Jun Qi)}}

\IEEEtitleabstractindextext{%
\begin{abstract}
As a representative self-supervised method, contrastive learning has achieved great successes in unsupervised training of representations.  It trains an encoder by distinguishing positive samples from negative ones given query anchors.  These positive and negative samples play critical roles in defining the objective to learn the discriminative encoder, avoiding it from learning trivial features.  While existing methods heuristically choose these samples, we present a principled method where both positive and negative samples are directly {\em learnable} end-to-end with the encoder.  We show that the positive and negative samples can be {\em cooperatively} and {\em adversarially} learned by {\em minimizing} and {\em maximizing} the contrastive loss, respectively. This yields cooperative positives and adversarial negatives with respect to the encoder, which are updated to continuously track the learned representation of the query anchors over mini-batches. The proposed method achieves $71.3\%$ and $75.3\%$ in top-1 accuracy respectively over $200$ and $800$ epochs of pre-training ResNet-50 backbone on ImageNet1K without tricks such as multi-crop or stronger augmentations. With Multi-Crop, it can be further boosted into $75.7\%$. The source code and pre-trained model are released in \url{https://github.com/maple-research-lab/caco}.
\end{abstract}

}

\maketitle

\IEEEdisplaynontitleabstractindextext

% \fi
%\IEEEpeerreviewmaketitle

\vspace{2.5em}
\IEEEraisesectionheading{\section{Introduction}\label{sec:introduction}}

Unsupervised representation learning \cite{oord2018representation,hjelm2018learning,zhang2019aet} has attracted tremendous attentions recently as model pre-training becomes an essential step for deep neural networks \cite{qi2019small}. Among them are a family of instance discrimination-based methods \cite{wu2018unsupervised,chen2020simple,he2019momentum,hu2021adco}, which train the network by distinguishing positive samples from their negative counterparts given query anchors from mini-batches
during the learning process. Contrastive learning \cite{oord2018representation} and many variants \cite{henaff2019data,tian2019contrastive} are one of the most popular directions that achieved great success in last 3 years by pulling the embeddings of positive pairs together and pushing that of negative pairs away. In these methods, a critical question arises regarding how to find the most critical positives and negatives that can provide discriminative knowledge to self-supervise the network training.

SimCLR \cite{chen2020simple} and MoCo \cite{he2019momentum,chen2020improved} are two representative contrastive learning methods in this category. Given a query anchor, SimCLR uses the other samples from the same mini-batch as negative examples while an augmented view of the anchor is used as the positive sample.  On the contrary, the MoCo maintains a memory bank as in \cite{wu2018unsupervised}, and uses the negative samples from the memory bank.  The memory bank is built by queueing the learned representations over past mini-batches in a FIFO (First-In-First-Out) fashion, and a momentum update is applied to the key network whose weights are an exponential moving average of the query networks over time. SimCLR \cite{chen2020simple} usually needs a very large batch size to involve sufficient negative samples in each iteration, making it very computationally and memory demanding. Although the MoCo also has a very large memory bank, it can be maintained efficiently as the negative samples in it are collected on-the-fly from the past mini-batches.

Both methods show that successful training strongly relies on the construction of negative samples. Usually, harder negatives that are closer and mixed with the query anchors contain more discriminative information, making it harder to learn shallow features providing shortcut solutions to distinguish between positives and negatives.  Along this direction, recent works \cite{kalantidis2020hard} use heuristic sampling by concentrating on the hard negatives around queries.

On the contrary, a recent breakthrough demonstrates that negative samples are directly learnable by treating them as a part of network weights through adversarial contrastive learning \cite{hu2021adco}, naturally resulting in hard negatives pushed towards query anchors. It provides a more principled way to learn rather than heuristically construct discriminative negatives. Specifically, by treating the negatives and the encoder as two adversarial players, the AdCo outperforms the SimCLR and MoCo in terms of both accuracy and efficiency.  However, a question still remains -- {\em can we learn positive samples as well just like those negative samples?} In this paper, we strive to answer it by revealing the relation between positive and negative samples via a cooperative-adversarial principle.

Unlike negative samples, we will show that given a query anchor, its positive sample ought to be learned {\em cooperatively} rather than adversarially with the encoder network. This is based on the observation that the representation of a positive sample that cooperatively minimizes the contrastive loss with the encoder will be pulled towards the given anchor, instead of being pushed away if it were adversarially learned by maximizing the loss. The cooperative positive sample will also be chosen from the memory bank shared with the negative samples.  This could mitigate false negative problem as memory bank samples in close proximity of a query anchor can be trained as positives, instead of being blindly treated as negatives. Both positive and negative samples will be end-to-end trained together with the encoder through such a Cooperative-adversarial Contrastive (CaCo) learning approach. Moreover, we propose to use the  most probable positive samples that are stable to update for an end-to-end training of the CaCo model. Experiment results demonstrate the proposed method achieves $71.3\%$ and $75.3\%$ in top-1 accuracy over 200 epochs and 800 epochs of pre-training on Imagenet1K without multi-crop augmentations. It can achieve 75.7\% in top-1 accuracy after multi-crop augmentations with 800 epochs pre-training.
%, outperforming the state-of-the-art methods that use multi-crop augmentations and/or larger batch sizes distributed across multiple GPU servers.

The remainder of the paper is organized as follows. We will review the related works in Section~\ref{sec:relawork}, and revisit the preliminary works on contrastive learning in Section~\ref{sec:prelim}. The proposed CaCo method is presented in Section~\ref{sec:CaCo}, followed by the experiments in Section~\ref{sec:exp}. We will conclude the paper in Section~\ref{sec:concl}.

\section{Related Work}
\label{sec:relawork}
In this section, we will review the related works on contrastive learning from three perspectives - instance discrimination-based contrastive learning, hard negative samples, and trainable memory bank.

{\bf\noindent Instance Discrimination.} Most of recent works on contrastive learning focus on instance discrimination-based methods \cite{wu2018unsupervised}. Given a query anchor from a training minibatch, it explicitly constructs the positive and negative samples, and self-trains the representation network (i.e., an encoder) by distinguishing positives from negatives. It minimizes the InfoNCE loss (a form of the contrastive loss) that can be implemented as a cross entropy by treating each instance as a positive or a negative class. The positive sample of a query anchor is often given as the augmented view of the same anchor.
Depending on different ways to form negatives, the contrastive learning has many variants.  Among them are the SimCLR \cite{chen2020simple} where negatives  come from the other samples of the same mini-batch, as well as the MoCo \cite{he2019momentum,chen2020improved} where exists a memory bank storing the representations over past mini-batches as negatives. The former usually relies on a large size of mini-batches to perform reliable training of the encoder, which is demanding for computing and memory. MoCo addresses this problem by reusing the obtained representation of past samples without having to re-computing them for each mini-batches. While some negative samples may belong to the same class as a query anchor in instance discrimination methods (i.e, false negatives), a nearest neighbor contrastive learning extends the SimCLR and the MoCo by instead assigning the negative most similar to the anchor as its positive \cite{dwibedi2021little}, which somehow mitigates the false negative problem.

{\bf\noindent Hard Negative Samples.} It has been observed that the harder negative samples that are difficult to distinguish from their positive counterparts play a more critical role in contrastive learning. They prevent the encoder from learning trivial features to shortcut the aforementioned instance discrimination task. There exist methods based on heuristically sampling the samples in proximity of positive anchors. For example, hard negative sampling methods \cite{robinson2020contrastive,kalantidis2020hard} apply importance sampling to find harder negatives with a concentration parameter. Recent works \cite{ho2020contrastive,kim2020adversarial} on adversarial pre-training also shed some light on learning representations resilient against adversarial attacks. However, they do not directly train negative samples end-to-end by treating memory bank as learnable network weights as in \cite{hu2021adco}. Instead, they apply the adversarial training to input images or augmented views that may adversely affect positive samples that ought to be constructed {\em cooperatively} as we find in this paper.

{\bf\noindent Trainable Memory Bank.} The recent work on Adversarial Contrastive learning (AdCo) \cite{hu2021adco} made a significant progress to demonstrate that the hard negatives are {\em directly learnable} as a part of network weights that are adversaries to the encoder network in the other part. Instead of minimizing the InfoNCE loss as the encoder, it maximizes it that pushes the learned negatives towards their query anchors. It provides a principled way to finding more discriminative negatives for an effective training of the encoder.  Along this line, an implicit feature modification (IFM) \cite{robinson2021can} is proposed to iteratively update the hard samples. The IFM also pushes the samples towards the representation of anchors like the AdCo, but its step size is set by a given level of budget. Instead, the iterative update direction of these samples in the AdCo is derived from the positive gradient of the InfoNCE loss as an adversary learner, and its step size is proportional to the probability of being a positive to a query anchor. In this way, the AdCo can produce more discriminative negatives by making it harder to distinguish them from the positive anchors.

In this paper, along the road outlined from instance discrimination and trainable memory bank, we will unify the joint training of both positive and negative samples given query anchors. Both positives and negatives will share a memory bank that is trained end-to-end. For this purpose, it needs to answer two critical questions in contrastive learning:
\begin{itemize}
\item Can we end-to-end train positive and negative samples jointly through a shared memory bank?
\item If so, will adversarial criterion still work for training positives?
\end{itemize}
The answer to these questions leads to the proposed Cooperative-adversarial Contrastive (CaCo) learning presented in this paper.
%These hard samples can avoid the trained encoder from learning trivial features.

%------------------------------------------------------------------------
\section{Preliminaries and Notations}\label{sec:prelim}
In this section, we revisit the contrastive learning and its adversarial training of memory bank samples. We will also define several notations that will be used later.

\subsection{Contrastive Learning}

Consider an unlabeled example $\mathbf x$ called a query anchor in the current training mini-batch and its representation $\mathbf z$ output from an encoder network $\mathbf z \triangleq \phi_\theta(\mathbf x)$ parameterized by $\theta$. To train the encoder $\phi_\theta$, a positive sample $\mathbf x^+$ of $\mathbf x$ is obtained by applying an augmentation (e.g., random crop, color jitting and Gaussian noises) to $\mathbf x$. The resultant $\mathbf x^+$ provides an alternate view of the original anchor $\mathbf x$, along with its representation $\mathbf z^+$. Usually, in a typical contrastive learning approach, the anchor $\mathbf x$ is also given by applying a random augmentation to the original image.

In instance discrimination-based contrastive learning such as MoCo~\cite{he2019momentum} and AdCo~\cite{hu2021adco}, together with mini-batches over iterations, a memory bank $\mathcal M \triangleq \{\mathbf b_j|j=1,\cdots,K\}$ of $K$ representations are also provided as negatives, in contrast to the positive representation $\mathbf z^+$ for an anchor. The memory bank is shared across all unlabeled anchors in a minibatch. Such a bank can be constructed by queueing the obtained representations from the past minibatches in a FIFO (First-In-First-Out) fashion such as in MoCo \cite{he2019momentum}. Alternatively, a more principled way is to directly learn the memory bank by maximizing the contrastive loss (i.e., InfoNCE loss), which has demonstrated more efficient to update the encoder \cite{hu2021adco}. In both cases, the encoder parameters $\theta$ are trained by 
$$
\theta^\star = \arg\min_{\theta} \sum_{\mathbf x\in\mathcal B}\ell(\mathbf x)
$$
over a mini-batch $\mathcal B$ with the following InfoNCE loss \cite{oord2018representation,wu2018unsupervised}
\begin{equation}\label{eq:infonce}
\ell(\mathbf x) = - \log \dfrac{\exp(\mathbf z\cdot \mathbf z^+/\tau)}{\exp(\mathbf z\cdot \mathbf z^+/\tau) + \sum_{j=1}^{K}{\exp(\mathbf z\cdot \mathbf b_j/\tau)}}
\end{equation}
where $\cdot$ denotes the inner product between two feature vectors and $\tau$ is the temperature controlling the sharpness of the loss. SimCLR \cite{chen2020improved} also symmetrizes the loss by switching $\mathbf z$ and $\mathbf z^+$ with the latter as a query anchor instead. Since the representations are normalized to a unit norm, the inner product gives the cosine similarity between samples.

\subsection{Adversarial Contrastive Learning}
Constructing the memory bank \cite{wu2018unsupervised,he2020momentum,chen2020improved} plays a critical role in contrastive learning.  The negative samples in it gives the important specification about how the underlying representation network $\phi_\theta$ ought to be learned by discriminating between positives and negatives. Previous results show that hard negatives could provide more useful information, since hard negative samples can prevent the encoder from learning trivial representations focusing on low-level features such as local details rather than high-level semantic structures such as object shapes and categories.

%the contrastive learning and its InfoNCE loss (\ref{eq:infonce}) trains the representation network to distinguish the positive representation $\mathbf z^+$ from the negative counterparts in the memory $\mathcal M$. Hard negative samples can prevent the encoder from learning shallow representations focusing on low-level features such as local details rather than high-level semantic representations such as object shapes and categories.

Furthermore, it has been shown that hard negative samples naturally result from an adversarial contrastive learning (AdCo) approach \cite{hu2021adco}, with a novel idea of directly learning the negatives to form a memory bank. This is contrary to the MoCo where the negative samples are collected by queueing the representations from past minibatches heuristically. Formally, the AdCo considers two adversary players -- the memory bank $\mathcal M$ and the representation encoder $\phi_\theta$. They are trained in a mutually adversarial manner by
\begin{equation}\label{eq:adco}
\mathcal M^\star, \theta^\star = \arg\max_{\mathbf b_j\in\mathcal M}\min_{\theta}\sum_{\mathbf x\in\mathcal B}\ell(\mathbf x)
\end{equation}
The previous results \cite{hu2021adco} show that the memory bank and the encoder can be updated alternately, resulting in more efficient representation learning since the negative samples can be updated more efficiently to provide critical information to specify the contrastive learning.

Although the AdCo \cite{hu2021adco} has made a significant progress to directly learn negative samples, the question still remains regarding if the positive sample $\mathbf x^+$ and its representation $\mathbf z^+$ for an unlabeled anchor can also be learned directly.  This will give rise to a more elegant solution by unifying the end-to-end training of both positives and negatives. Meanwhile, it could also mitigate the false negative problem in contrastive learning, since it is known that some negative samples in the learned memory bank would be positive to an anchor if their representations are sufficiently close. This inspires us to develop the following Cooperative-adversarial Contrastive (CaCo) learning method.

\begin{figure}[t]
%\captionsetup[subfigure]{singlelinecheck=false}
    \centering
    \begin{minipage}[c]{\linewidth}
    \begin{center}
        \subfloat[MoCo]{\includegraphics[width=\textwidth]{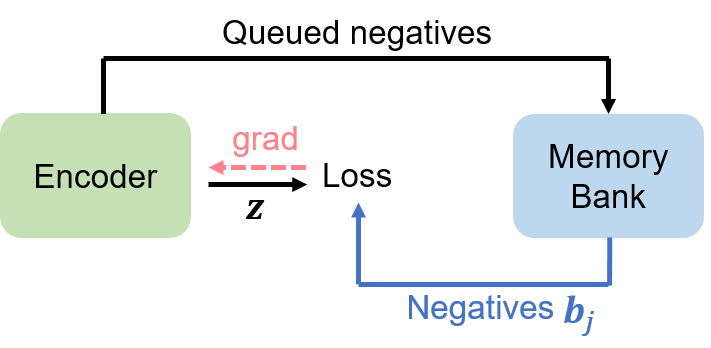}}
        
    \end{center}
    \end{minipage}
    \begin{minipage}[c]{\linewidth}
    \begin{center}
        \subfloat[AdCo]{\includegraphics[width=\textwidth]{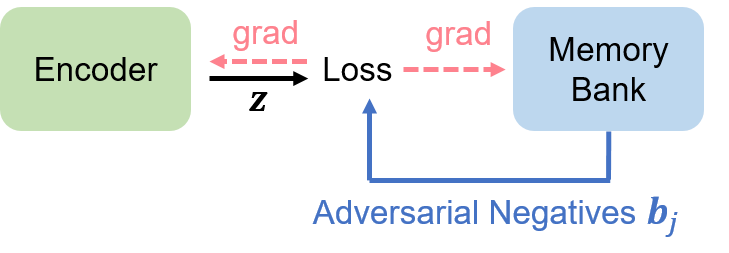}
        }
    \end{center}
    \end{minipage}
    \begin{minipage}[c]{\linewidth}
    \begin{center}
        \subfloat[CaCo]{\includegraphics[width=\textwidth]{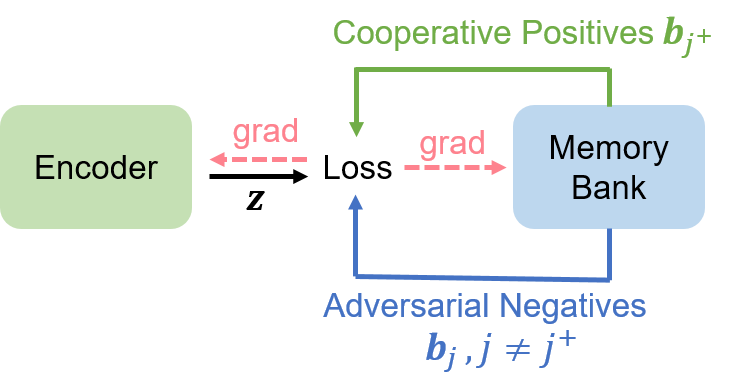}}
        
    \end{center}
    \end{minipage}
    \caption{This figure compares existing contrastive learning methods -- MoCo and AdCo -- with the proposed CaCo. Both MoCo and AdCo only has negative samples from the memory bank, and MoCo queues these samples from the encoder output over past minibatches without training them directly. In contrast, given a query anchor $\mathbf z$, CaCo directly learns the cooperative positive $\mathbf b_{j^+}$ and adversarial negatives $\mathbf b_{j}, j\neq j^+$ from the shared memory bank through backward gradients (${\rm grad}$).
    %By comparison, MoCo trains neither negatives nor positives, while AdCo only trains adversarial negatives.
    }\label{fig:comparison}
\end{figure}

Figure~\ref{fig:comparison} gives a glance at the differences between MoCo \cite{he2019momentum,chen2020improved}, AdCo \cite{hu2021adco} and the proposed CaCo. Both MoCo and AdCo only has negative samples from the memory bank, and MoCo queues these samples from the past mini-batches without training them directly. In contrast, both positive and negative samples are obtained from the shared memory bank given a query anchor $\mathbf z$, and they are directly trained through backward gradients by minimizing and maximizing the loss, respectively. The proposed CaCo will be presented in detail below.

\section{CaCo: Cooperative-adversarial Contrastive Learning}\label{sec:CaCo}
In this section, we will present the proposed Cooperative-adversarial Contrastive (CaCo) Learning. First we will explicitly integrate positive samples shared with negatives from the memory bank  into the contrastive learning in Section~\ref{sec:loss}. Then in Section~\ref{sec:cpan} we will show that positives and negatives ought to play cooperative and adversarial role in the network training, respectively, followed by the elaboration on cooperative-adversarial training over minibatches in Section~\ref{sec:caco_training}.
%We will discuss how to choose the positive samples for anchor examples in Section~\ref{sec:mcp}.

\subsection{Positives from Shared Memory Bank}\label{sec:loss}
Following the definition defined in the last section, consider the representation $\mathbf z$ of an unlabeled anchor $\mathbf x$. In the AdCo, we assume its negative samples all come from the trainable memory bank $\mathcal M$. Now let us make a bold step by assuming the corresponding positive representation also comes from $\mathcal M$.  We denote the index of the positive sample in $\mathcal M$ by $j^+\in\mathcal K\triangleq  1,\cdots,K$. Thus, for $\mathbf x$ with embedding $z$, its positive representation is given by $\mathbf b_{j^+}\in\mathcal M$. Accordingly, the contrastive InfoNCE loss becomes
\begin{equation}\label{eq:caco}
\begin{aligned}
\ell(\mathbf x) &\triangleq  - \log \dfrac{\exp(\mathbf z\cdot \mathbf b_{j^+}/\tau)}{\exp(\mathbf z\cdot \mathbf b_{j^+}/\tau) + \sum_{j\neq j^+}{\exp(\mathbf z\cdot \mathbf b_j/\tau)}}\\
&=- \log \dfrac{\exp(\mathbf z\cdot \mathbf b_{j^+}/\tau)}{\sum_{j=1}^K{\exp(\mathbf z\cdot \mathbf b_j/\tau)}}
\end{aligned}
\end{equation}
We note that since both the positive and negative samples come from the shared memory bank, all terms in the denominator can be written uniformly with the same set of the memory bank representations, yielding a more elegant contrastive loss.

\subsection{Cooperative Positives vs. Adversarial \\ Negatives}\label{sec:cpan}
An unlabeled anchor $\mathbf x$ has both positive and negative samples, which, as aforementioned, we assume are all directly learnable. Following the gradient decent method, we can derive the gradient of (\ref{eq:caco}) over the positive sample $\mathbf b_{j^+}$ and its negative counterparts $\mathbf b_j$ for $j\neq j^+$ to update them.

It is not hard to obtain the following gradients
\begin{equation}\label{eq:pos_grad}
\dfrac{\partial \ell(\mathbf x)}{\partial \mathbf b_{j^+}} = - \dfrac{1}{\tau} \left[1-p(\mathbf b_{j^+}|\mathbf z)\right] \mathbf z
\end{equation}
and
$$
\dfrac{\partial \ell(\mathbf x)}{\partial \mathbf b_{j}} =  \dfrac{1}{\tau} p(\mathbf b_{j}|\mathbf z) \mathbf z,~j\neq j^+
$$
which are for the positive and negative samples, respectively. Here, $p(\mathbf b_{j}|\mathbf z)$ denotes the probability of a memory bank sample $b_j$ being positive to the input representation $\mathbf z$,
\begin{equation}\label{eq:prob_positive}
p(\mathbf b_{j}|\mathbf z) = \dfrac{\exp(\mathbf z\cdot \mathbf b_{j}/\tau)}{\sum_{k=1}^K{\exp(\mathbf z\cdot \mathbf b_k/\tau)}}
\end{equation}

Based on the idea behind the adversarial training, the negative samples can thus be updated by maximizing $\ell(\mathbf x)$, resulting in
$$
\mathbf b_j \leftarrow \mathbf b_j + \dfrac{\eta}{\tau} p(\mathbf b_{j}|\mathbf z) \mathbf z,~j\neq j^+,
$$
where $\eta$ is the learning rate for the gradient update. This shows that each negative representation $\mathbf b_j$ will be updated {\bf towards} the anchor representation $\mathbf z$, which will make it harder to distinguish the negative sample from the anchor $\mathbf z$. The previous results show that such harder negatives can provide more useful information to train the encoder. In this sense, the negative samples are {\em adversarial} players to the encoder. For the adversarial negative example training, AdCo~\cite{hu2021adco} has carefully discussed it and utilized it for contrastive learning successfully. 

Now the question is: {\em should we also make the positive sample an adversary to the encoder}? Let us take a look at the gradient of the positive representation $\mathbf b^+$ in (\ref{eq:pos_grad}). If it were assume to be an adversary, it would be updated by
$$
\mathbf b_{j^+} \leftarrow \mathbf b_{j^+} {\color{red}\boldsymbol -} \dfrac{\eta}{\tau} \left[1-p(\mathbf b_{j^+}|\mathbf z)\right] \mathbf z
$$
In this case, the positive $\mathbf b_{j^+}$ would be pushed {\bf away} from the anchor $\mathbf z$, instead of being pushed towards it.  This causes a dilemma -- what we chose as a positive sample would eventually end up being pushed away from the anchor, making it more likely to be a negative rather than a positive sample. Obviously, this is not our intention to learn a positive sample. 

This shows that the positive sample should be a {\bf cooperative} player rather than an adversary to the encoder by minimizing the contrastive loss. This results in the following update to the positive sample
$$
\mathbf b_{j^+} \leftarrow \mathbf b_{j^+} {\color{red}\boldsymbol +} \dfrac{\eta}{\tau} \left[1-p(\mathbf b_{j^+}|\mathbf z)\right] \mathbf z
$$
which in turn pushes the positive representation towards the anchor, in line with our intuition to learn the positive sample close to the anchor.

\begin{figure}[t]
%\captionsetup[subfigure]{singlelinecheck=false}
    \centering
     \includegraphics[width=0.47\textwidth]{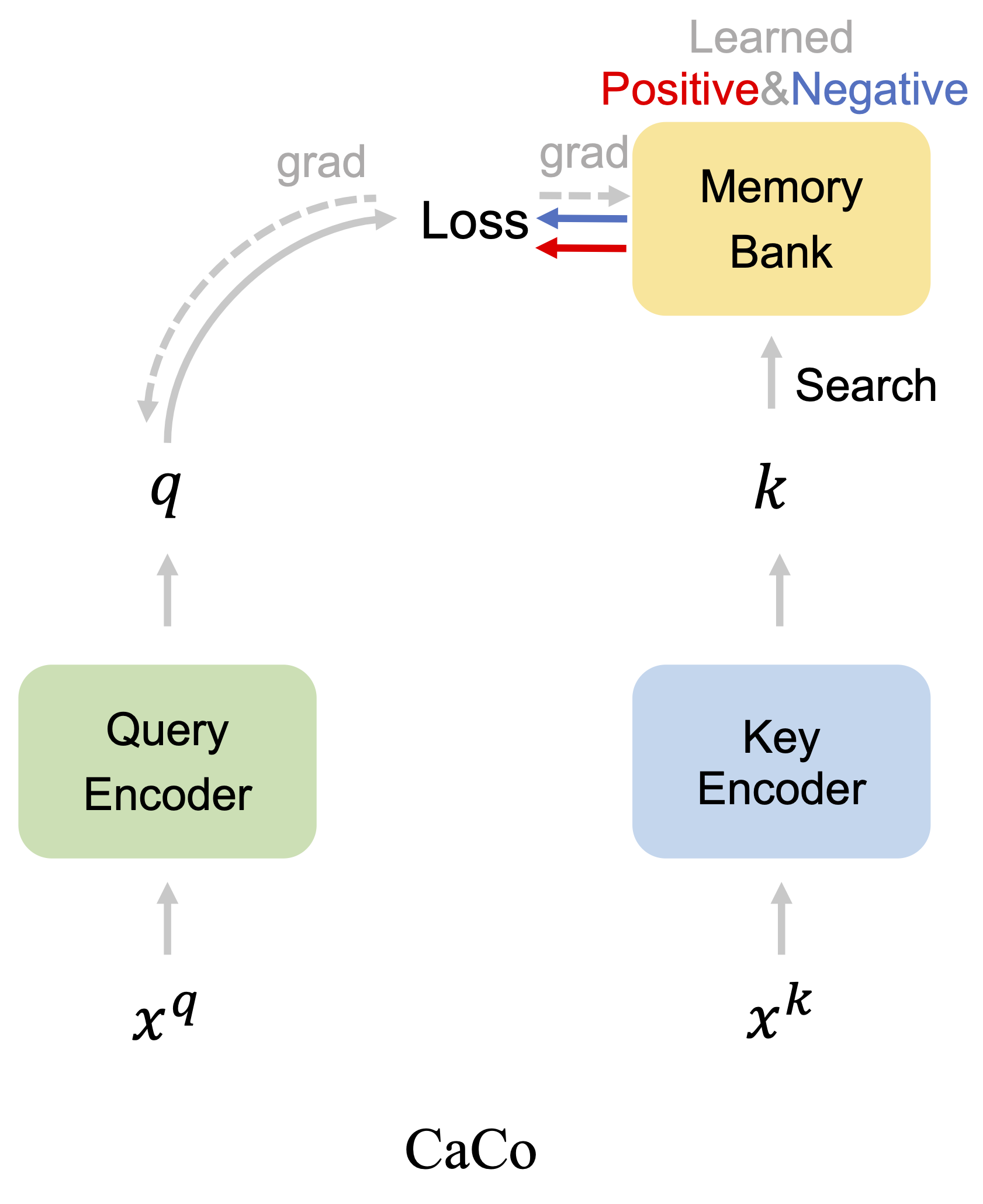}
     
    \caption{The diagram of CaCo. CaCo trains a visual representation encoder and a cooperative-adversarial memory bank at the same time. For a encoded query $q$, we used its counterpart key $k$ by the key encoder to identify its most probable positive out of memory bank as its positive pair, while all other embeddings in the memory bank are its negative pairs. Then the network is optimized by minimizing the contrastive loss, the memory bank is coopertive-adversarial optimized as discussed in the paper.}
    %By comparison, MoCo trains neither negatives nor positives, while AdCo only trains adversarial negatives.
    \label{fig:diagram}

\end{figure}

%Figure~\ref{fig:comparison}(c) illustrates the CaCo directly learns the cooperative positive $\mathbf b_{j^+}$ and adversarial negatives $\mathbf b_{j}, j\neq j^+$ from memory bank given a query anchor $\mathbf z$. They are learned through backward gradients by minimizing and maximizing the loss, respectively.
%In contrast, MoCo trains neither negatives nor positives, while AdCo only trains adversarial negatives.

\subsection{Cooperative-Adversarial Training}\label{sec:caco_training}
The above analysis yields the following Cooperative-Adversarial Contrastive (CaCo) objective
\begin{equation}\label{eq:b_caco}
\sum_{\mathbf z\in\mathcal B} \min_{\mathbf b_{j^+},\theta}~\max_{\mathbf b_j,j\neq j^+} - \log \dfrac{\exp(\mathbf z\cdot \mathbf b_{j^+}/\tau)}{\sum_{j=1}^K{\exp(\mathbf z\cdot \mathbf b_j/\tau)}}
\end{equation}
where both the positive sample $\mathbf b_{j^+}$ and the encoder with the parameters $\theta$ minimizes the loss, thus being cooperative players, while the negative samples $\mathbf b_j, j\neq j^+$ are adversary instead to maximize it.

The above CaCo objective is defined over a mini-batch $\mathcal B$. One can assign a distinct positive sample from the memory bank to each anchor $\mathbf z$ in $\mathcal B$. In other words, the positive index $j^+$ is a function of $\mathbf z$ in (\ref{eq:b_caco}). Thus, for a memory bank sample $\mathbf b_j$, we use $\mathcal P_j^+$ ($\mathcal N_j^-$ resp.) $\subseteq \mathcal B$ to denote all anchors $\mathbf z$'s, for which $\mathbf b_j$ is a positive (negative resp.) sample.
For each $\mathbf b_j$, we have $\mathcal P_j^+ \bigcap \mathcal N_j^+ = \emptyset$ and $\mathcal P_j^+ \bigcup \mathcal N_j^+ = \mathcal B$. Then, according to (\ref{eq:b_caco}), all memory bank samples can be updated by
\begin{equation}\label{eq:update}
\mathbf b_j \leftarrow \mathbf b_j {\color{blue}\boldsymbol-} \sum_{\mathbf z \in\mathcal P_j^+}\eta \dfrac{\partial \ell(\mathbf x)}{\partial b_j^+} {\color{red}\boldsymbol+} \sum_{\mathbf z \in \mathcal N_j^-} \eta \dfrac{\partial \ell(\mathbf x)}{\partial b_j}
\end{equation}
with a positive learning rate $\eta$, where the minus ``${\color{blue}\boldsymbol-}$" and the plus ``${\color{red}\boldsymbol+}$" in front of the last two terms reflect the cooperative and the adversarial updates to $\mathbf b_j$. %resulting respectively from the positives and negatives.

\begin{figure*}[t]
\small
%\captionsetup[subfigure]{singlelinecheck=false}
    \centering
    \includegraphics[width=0.65\textwidth]{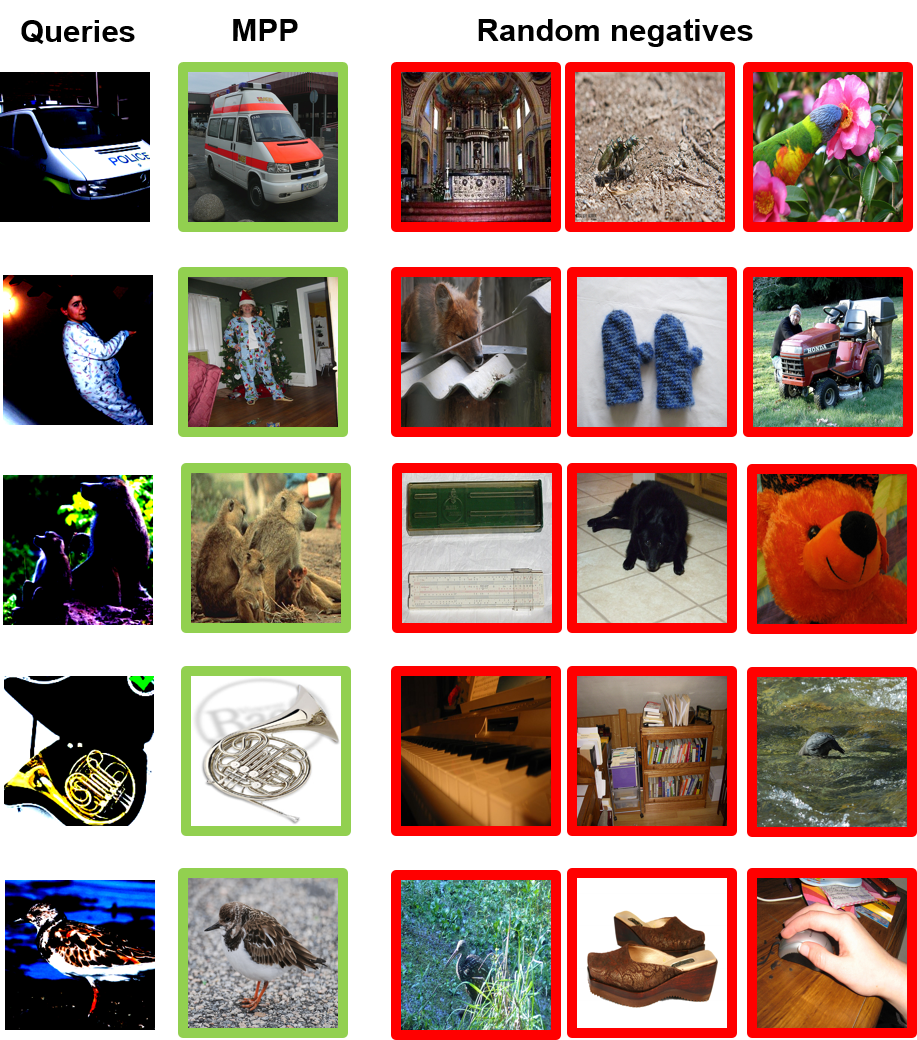}
    \caption{The surrogate images of the Most Probable Positive (MPP) and random negatives from the learned memory bank for some query examples.}
    %By comparison, MoCo trains neither negatives nor positives, while AdCo only trains adversarial negatives.
    \label{fig:query_ex}
\end{figure*}

{\noindent\bf Effect of $\l_2$-normalization on memory bank updates.} It is worth noting that the gradient in (\ref{eq:pos_grad}) was derived by assuming both $\mathbf z$ and all memory bank samples $\mathbf b_j$'s have already been normalized to a unit length. Even though they are normalized, the gradient update will result in new vectors not normalized to a unit length anymore.  Thus, an $l_2$-normalization is often required after the gradient update. Such a normalization can be built into (\ref{eq:caco}) by replacing $\mathbf z$ and $\mathbf b_j$ with $\mathbf z/\|\mathbf z\|_2$ and $\mathbf b_j/\|\mathbf b_j\|_2$, respectively. This results in the following gradient
$$
\dfrac{\partial \ell(\mathbf x)}{\partial \mathbf b_{j^+}} = - \dfrac{1}{\tau} \left[1-p(\mathbf b_{j^+}|\mathbf z)\right]\dfrac{\|\mathbf b_{j^+}\|^2_2\mathbf z - \mathbf z \cdot \mathbf b_{j^+} \mathbf b_{j^+}}{\|\mathbf z\|_2\|\mathbf b_{j^+}\|^3_2}
$$
This shows how to update the positive sample end-to-end by viewing its representation $\mathbf b_{j^+}$ as free network weights. For simplicity, one can assume every time after a mini-batch update, $\mathbf b_{j^+}$ and $\mathbf z$ are normalized back to a unit length such that we always keep the current samples $\|\mathbf b_{j^+}\|_2=1$ and $\|\mathbf z\|_2=1$. In this case, the above gradient can be simplified to
\begin{equation}\label{eq:nor_pos_grad}
\begin{aligned}
\dfrac{\partial \ell(\mathbf x)}{\partial \mathbf b_{j^+}} &= - \dfrac{1}{\tau} \left[1-p(\mathbf b_{j^+}|\mathbf z)\right](\mathbf z - \mathbf z \cdot \mathbf b_{j^+} \mathbf b_{j^+}) \\
&= - \dfrac{1}{\tau} \left[1-p(\mathbf b_{j^+}|\mathbf z)\right](\mathbf I -  \mathbf b_{j^+} \mathbf b_{j^+}^T)\mathbf z
\end{aligned}
\end{equation}

Similarly, we can derive the gradient over negative samples by assuming all $\mathbf b_j$'s are normalized to a unit norm after each gradient update
\begin{equation}\label{eq:nor_neg_grad}
\begin{aligned}
\dfrac{\partial \ell(\mathbf x)}{\partial \mathbf b_{j}} &=  \dfrac{1}{\tau} p(\mathbf b_{j}|\mathbf z)(\mathbf I -  \mathbf b_{j} \mathbf b_{j}^T)\mathbf z, j\neq j^+
\end{aligned}
\end{equation}
From (\ref{eq:nor_pos_grad}) and (\ref{eq:nor_neg_grad}), we can see that both positive and negative samples will be updated towards the transformed direction $\mathbf T_j \mathbf z$ with $\mathbf T_j = \mathbf I - \mathbf b_j\mathbf b_j^T$. It is easy to verify that
\begin{itemize}
\item The gradient $\mathbf T_j \mathbf z$ is orthogonal to $\mathbf b_j$, i.e., $\mathbf b_j^T \mathbf T_j \mathbf z = 0$. It suggests that the gradient is tangent to the unit-hypersphere such that it tends to keep the unit norm of $\mathbf b_j$ unchanged as expected.
\item Along the gradient, the sample $\mathbf b_j$ also tends to be pushed closer towards $\mathbf z$ since their cosine similarity $\mathbf z^T \mathbf T_j \mathbf z = 1-(\mathbf z \cdot \mathbf b_j)^2 \geq 0$ is nonnegative.
\end{itemize}

\begin{table}
\small
\begin{center}
\setlength{\tabcolsep}{3.5mm}{
\begin{tabular}{lccc}
\toprule[1pt]
Method & batch size&top-1 \\
\midrule[1pt]
InstDisc~\cite{wu2018unsupervised} & 256  & 58.5 \\
SimCLR~\cite{chen2020simple} & 256  & 61.9 \\
CPC v2~\cite{henaff2019data} & 512  & 63.8 \\
PCL v2~\cite{li2020prototypical} & 256  & 67.6 \\
MoCo v2~\cite{chen2020improved} & 256  & 67.5 \\
PIC~\cite{cao2020parametric} & 512 & 67.6 \\
InfoMin Aug~\cite{tian2020makes} & 256  & 70.1\\
SimSiam~\cite{chen2020exploring} &256&70.0\\
SwAV~\cite{caron2020unsupervised} & 4096  & 69.1 \\
NNCLR~\cite{dwibedi2021little} &4096&70.7\\
BYOL~\cite{grill2020bootstrap} & 4096 &70.6\\
AdCo~\cite{hu2021adco} &256  & 68.2 \\
%NNCLR~\cite{dwibedi2021little} &1024& xx \\
\midrule
CaCo &256 &{\bf 70.9}\\
CaCo &1024 &{\bf 71.3}\\
CaCo &4096 &{\bf 72.0}\\
\bottomrule[1pt]
\end{tabular}}
\end{center}

\caption{\small Top-1 accuracy under the linear evaluation on Imagenet1K dataset with the ResNet-50 backbone. All compared methods use single-crop augmentations pre-trained over $200$ epochs.}
\label{tab:imagenet200}

\end{table}
Now, we can rewrite the gradient update (\ref{eq:update}) to all samples $j=1,\cdots,K$ in the memory bank as
\begin{equation}\label{eq:real_update}
\begin{aligned}
\mathbf b_j \leftarrow \mathbf b_j & \underbrace{+ \sum_{\mathbf z \in\mathcal P_j^+}\dfrac{\eta}{\tau} \left[1-p(\mathbf b_{j}|\mathbf z)\right] \mathbf T_j\mathbf z}_{{\rm cooperative~~positives}}\\
&\underbrace{+ \sum_{\mathbf z \in \mathcal N_j^-} \dfrac{\eta}{\tau} p(\mathbf b_{j}|\mathbf z) \mathbf T_j\mathbf z}_{{\rm adversarial~~negatives}}.
\end{aligned}
\end{equation}
From this update, we observe that
\begin{itemize}
\item For cooperative positives, the more probable $\mathbf b_j$ is positive to $\mathbf z$ with a higher $p(\mathbf b_j|\mathbf z)$, the less it will be pushed towards $\mathbf z$. This is intuitive as a positive sample already closer to its anchor $\mathbf z$ does not need to be updated too much;
\item For adversarial negatives, we have an opposite observation -- the more probable $\mathbf b_j$ is positive to $\mathbf z$, the more it should be pushed towards the anchor. In this way, it will become a harder negative to be distinguished from the positive counterparts, thereby providing more discriminative information to update the encoder.
\end{itemize}

\begin{table}
\small
\begin{center}
\setlength{\tabcolsep}{3.5mm}{
\begin{tabular}{lccc}
\toprule[1pt]
Method & epochs & batch size&top-1 \\
\midrule[1pt]
%Supervised &-&256& 76.5\\
%\hline
%BigBiGAN~\cite{donahue2019large}& - & 2048 & 56.6 \\
SeLa~\cite{asano2019self} &400& 256  & 61.5 \\
PIRL~\cite{misra2020self}&800& 1024 & 63.6 \\
CMC~\cite{tian2019contrastive} &240& 128  & 66.2\\
SimCLR~\cite{chen2020simple} &800& 4096 & 69.3\\
PIC~\cite{cao2020parametric} &1600& 512 &  70.8 \\
MoCo v2~\cite{chen2020improved} &800& 256 &  71.1 \\
InfoMin Aug~\cite{tian2020makes}&800& 256 &  73.0\\
SimSiam~\cite{chen2020exploring} &800&256&71.3\\
SwAV~\cite{caron2020unsupervised} &800& 4096  & 71.8 \\
NNCLR~\cite{dwibedi2021little} &800&1024&72.9\\
BYOL~\cite{grill2020bootstrap}&1000& 4096 & 74.3\\
AdCo~\cite{hu2021adco}&800 &256  & 72.8 \\
MoCo v3~\cite{chen2021empirical}&1000&4096&74.6\\
NNCLR~\cite{dwibedi2021little} &800&4096&74.9\\
\midrule
CaCo &800&1024 & {\bf 74.1}\\
CaCo &800&4096& {\bf 75.3}\\
\bottomrule[1pt]
\end{tabular}}
\end{center}

\caption{\small Top-1 accuracy under the linear evaluation on Imagenet1K with the ResNet-50 backbone. The table compares the methods using a single crop augmentation pre-trained with more epochs.}
\label{tab:imagenet800}

\end{table}

%\subsubsection*{The Most Probable Positive}\label{sec:mcp}
{\noindent \bf The Most Probable Positive.} The last question we need to answer is how we choose the positive sample given an anchor. %Obviously, the most natural choice is the most probable positive sample.
Here, we propose to use the {\em Most Probable Positive} (MPP) sample resulting in the minimum change that favors a stable update for the training purpose.

Specifically, once a positive sample $\mathbf b_{j^+}$ is chosen for an anchor $\mathbf z$, its representation will be updated along the (negative) gradient based on (\ref{eq:nor_pos_grad}).  Ideally, we expect that the updated sample will still be positive to the anchor.  This ensures the positive assignments in $\mathcal P^+_j$ (and thus $\mathcal N^-_j$ as well) be as stable as possible so that they can be updated stably based on (\ref{eq:real_update}).

For this reason, we note that $[1-p(\mathbf b_{j}|\mathbf z)]$ is the step size (up to a constant coefficient $\frac{1}{\tau}$) to update the positive sample $\mathbf b_{j^+}$ in (\ref{eq:nor_pos_grad}). To minimize the change to the positive sample, we choose the positive sample that minimizes this step size,
\begin{equation}\label{eq:max_prob}
j^+ = \arg\min_j 1- p(\mathbf b_{j}|\mathbf z) = \arg\max_j p(\mathbf b_{j}|\mathbf z),
\end{equation}
which results in the Most Probable Positive (MPP) sample to the anchor $\mathbf z$. Intuitively, the chosen positive $\mathbf b_{j^+}$ will be the one that minimizes the error in assigning a false positive sample to $\mathbf z$ based on $p(\mathbf b_{j}|\mathbf z)$ in (\ref{eq:prob_positive}).

To address the most probable positive clearly, the diagram of the training process is illustrated in Fig.~\ref{fig:diagram}. For a fair comparison with the SOTA methods such as MoCo~\cite{he2019momentum} and AdCo~\cite{hu2021adco}, the CaCo also has two streams composed of a momentum encoder and a query encoder, where the momentum encoder is used to identify the positives in the memory bank, and the contrastive loss will be propagated to query encoder and memory bank to train them. All positive and negative samples are treated as free parameters like the network weights, and they are directly learned through gradient descent/ascent by back-propagating the contrastive loss in the proposed cooperative-adversarialfashion, as shown in Eqn.~(\ref{eq:real_update}).

{\noindent \bf Avoidance of collapse.} To explain the avoidance of collapse, it is worth noting that both positive and negative samples share the memory bank across query anchors in a mini-batch, and they are dynamically updated over iterations in the proposed cooperative-adversarial training. In other words, a memory bank sample $b$ can be positive for one query anchor, while being negative for another one, which keeps the memory bank from collapsing to a single trivial point. Memory bank samples will also be updated over iterations, and  different positives will be assigned to the feature representation $z$ of a sample, keeping the latter from collapsing to a prefixed point $b^+$.

\section{Experiments}\label{sec:exp}
In this section, we evaluate the proposed CaCo and compare the results with the other unsupervised models including the state-of-the-art contrastive learning methods.

\begin{table}
\small
\begin{center}
\setlength{\tabcolsep}{3.5mm}{
\begin{tabular}{lccc}
\toprule[1pt]
Method & epochs & batch size&top-1 \\
\midrule[1pt]
SwAV~\cite{caron2020unsupervised} &800& 4096  & 75.3 \\
DINO~\cite{caron2021emerging} &800&4096&75.3\\
NNCLR~\cite{dwibedi2021little} &800&4096&75.6\\
\midrule
CaCo &800&2048& {\bf 75.7}\\
\bottomrule[1pt]
\end{tabular}}
\end{center}

\caption{\small Top-1 accuracy under the linear evaluation on Imagenet1K with the ResNet-50 backbone. The table compares the methods using same multi-crop augmentation in SwAV~\cite{caron2020unsupervised} pre-trained with more epochs.}
\label{tab:imagenetmulti}

\end{table}
%##################################################################################################
\begin{table}
\begin{center}

\setlength{\tabcolsep}{1mm}{
\begin{tabular}{llclc}
\toprule[1pt]
Method &Epoch&Time(h)& GPU & \begin{tabular}[c]{@{}l@{}}GPU Time\\ /epoch\end{tabular}\\
%Total GPU time/Epoch(h) \\ 
\midrule[1pt]
MoCo v2\cite{chen2020improved} &200 & 53.0 & 8$\times$V100 & 2.12\\
BYOL\cite{grill2020bootstrap}& 1000 & 8.0 & 512$\times$TPU & 4.10 \\
SWAV*\cite{caron2020unsupervised} & 800 & 50 & 64$\times$V100 & 4.06 \\
AdCo~\cite{hu2021adco} &200&56.5& 8$\times$V100 & 2.26 \\
\midrule
CaCo  &200&56& 8$\times$V100 & 2.24\\
\bottomrule[1pt]
\end{tabular}}
\end{center}
\caption{Running time comparison of different methods. GPU time/Epoch means sum of time cost of all GPUs to train an epoch. Here the time are measured under the asymmetrical settings to have fair comparison with early methods}
\label{tab:time_cost}
\end{table}
%##################################################################################################

\begin{table*}
\small
\begin{center}
\resizebox{\textwidth}{!}
{
\begin{tabular}{lccccccccccc}
\toprule[1pt]
Method  & SUN397 & Cars &Food101 & Aircraft & VOC2007 &  DTD & Pets & Caltech-101 & Flowers&  CIFAR10 & CIFAR100\\
\midrule[1pt]
% Supervised &-&256& 76.5\\
% \hline
SimCLR~\cite{chen2020simple} & 60.6& 49.3&72.8& 49.8& 81.4&75.7&84.6&89.3&92.6& 90.5 & 74.4 \\
MoCo v2~$^\dag$~\cite{chen2020improved} &61.1&67.0&71.5&50.9 & 87.1&74.5&87.6&91.4&90.0 & 90.9 &  74.8\\
AdCo~$^\ddag$~\cite{hu2021adco} & 62.2 & 66.8 &73.8 & 63.6&92.4 & 75.5& 86.8& 91.8& 93.7&91.2  & 74.3 \\
BYOL~\cite{grill2020bootstrap} & 62.2& 67.8&75.3& 60.6 &82.5 &75.5 &90.4 &94.2&{96.1}& 91.3 & 78.4\\
NNCLR~\cite{dwibedi2021little} & 62.5 &67.1&{76.7}&64.1&83.0&75.5&91.8&91.3&95.1&{93.7}&{79.0}\\
\midrule
CaCo &\textbf{64.6}&\textbf{73.8} &75.4&\textbf{66.0}& \textbf{92.6}& \textbf{76.8} & \textbf{91.9}&\textbf{94.4}& \textbf{96.1}&92.6 &76.0  \\
\bottomrule[1pt]
\end{tabular}
}
% }
\end{center}

\caption{ Transfer learning results on various datasets with ResNet-50 pretrained with Imagenet1K over $800$ epochs. The results are obtained with $^\dag$ \url{https://github.com/facebookresearch/moco} under the CC-BY-NC 4.0 license and $^\ddag$ \url{https://github.com/maple-research-lab/AdCo/} under MIT license. Other results were reported directly in the other papers.}
\label{tab:transfer}

\end{table*}

\subsection{Implementation Details}
\begin{figure}[t]
%\captionsetup[subfigure]{singlelinecheck=false}
    \centering
    \includegraphics[width=0.5\textwidth]{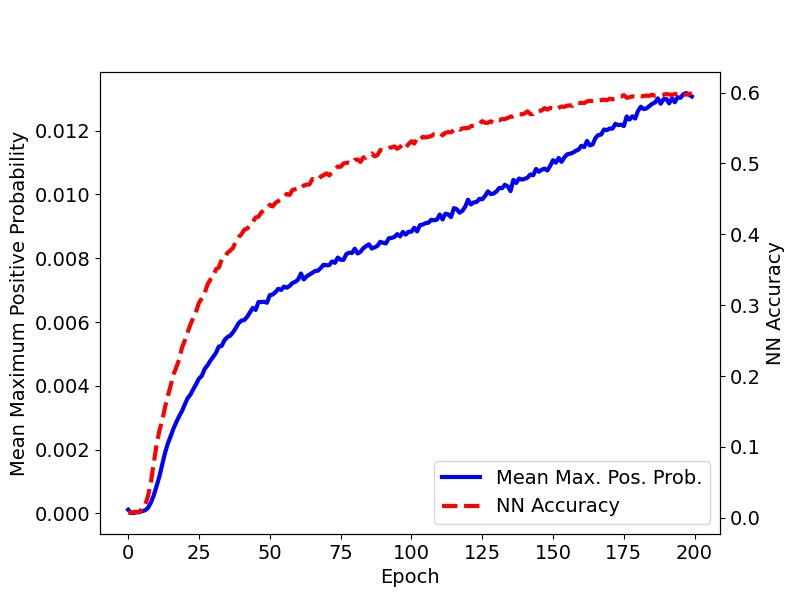}
    \caption{Plot of Mean Maximum Positive Probability and NN accuracy (based on $20\%$ training set) over $200$ epochs of pretraining on Imagenet1K.}
    %By comparison, MoCo trains neither negatives nor positives, while AdCo only trains adversarial negatives.
    \label{fig:prob_acc}
    
\end{figure}

For a fair comparison with the existing models \cite{he2019momentum,chen2020simple}, we adopt the ResNet-50 as the backbone for unsupervised pre-training on ImageNet. The output feature map from the top ResNet-50 block is average-pooled and projects to a 256-D feature vector through three MLP layers with two 2048-D hidden layers and the ReLU \cite{chen2020simple}. The resultant vector is $\ell_2$ normalized to calculate the cosine similarity. We adopt the single-crop protocol used in \cite{chen2020simple,he2019momentum,chen2020improved} to augment data in single-crop experiments and we used multi-crop augmentations used in literature \cite{caron2020unsupervised} to apply multi-crop in final experiments that can further boost performances.

We adopt up to $4,096$ images in a mini-batch for pre-training across up to four 8-Nvidia Tesla V100 GPU servers for single-crop. For the multi-crop setting, based on the same number of four 8-Nvidia Tesla V100 GPU servers, we adopt a batch size of $2,048$ images for pre-training.
%as it is the largest size that can be accommodated in a single GPU server with eight V100 cards. 
We will show that the proposed CaCo could also achieve competitive results with smaller batch sizes.
 %When multiple GPU servers are used, the batch size will be multiplied by the number of servers by convention.
By convention, the number of negative adversaries is set to $65,536$ to make a fair comparison with the other methods \cite{he2019momentum,hu2021adco}.

For the network pre-training, we adopt an initial learning rate of $0.03*\rm{batch\_size}/256$ and  $0.3*\rm{batch\_size}/256$ in the SGD \cite{bottou2010large} and LARS optimizer \cite{you2017large} for updating the backbone network for smaller ($<1024$) and larger ($\geq 1024$) batch sizes, respectively, along with a weight decay of $10^{-4}$ and a momentum of $0.9$. The memory bank is always directly optimized by SGD optimizer with learning rate $3.0$ without weight decay and a momentum of $0.9$. The cosine scheduler \cite{loshchilov2016sgdr} is used to gradually decay the learning rates. We set the same temperate $\tau$ to $0.08$ when updating the memory bank and network backbone.
%This differs from AdCo where two separate temperatures were adopted. 
After the backbone network is initialized, its output feature vectors over randomly drawn training images are used to initialize the memory bank. After that, the encoder and the shared memory bank with both positives and negatives are alternately updated.

%update the negative adversaries than that ($\tau=0.12$) used for updating the backbone network. This makes the updated negative adversaries sharper in distinguishing themselves from the positive images, and thus they will be nontrivially challenging examples in training the network.

%We also set the temperatures for updating the backbone network and the negative adversaries to $0.1$ and $0.02$, respectively.
\begin{table*}
\begin{center}

\begin{tabular}{ccccccc}
\toprule[1pt]
Method & Symmetric Loss & Batch Size & \#Negative Samples & \#Params. & Top-1 & \begin{tabular}[c]{@{}l@{}}(GPU $\cdot$ Time)\\ /epoch\end{tabular}\\ 
\midrule[1pt]
SimCLR~\cite{chen2020simple} & \checkmark & 8192 & - & - & 67.0 & 1.92 \\
MoCo v2~\cite{chen2020improved} & \checkmark & 256 & 65536 & 8M & 70.2 & 3.34 \\
BYOL~\cite{grill2020bootstrap} & \checkmark & 4096 & - & 1M & 70.6 & 4.10 \\
SimSiam~\cite{chen2020exploring} & \checkmark & 256 & - & 2M & 70.0 & - \\
AdCo~\cite{hu2021adco} & \checkmark & 256 & 65536 & 8M & 70.5 & 3.50 \\
AdCo~\cite{hu2021adco} & \checkmark & 256 & 16384 & 2M & 70.0 & 3.46 \\
AdCo~\cite{hu2021adco} & \checkmark & 256 & 8192 & 1M & 70.2 & 3.45 \\
\midrule
CaCo  & \checkmark & 4096 & 37268&4M&71.7& 3.83 \\
CaCo  & \checkmark & 4096 & 65536&8M&72.0& 3.84 \\
CaCo  & \checkmark & 4096 & 98304&12M&72.1& 3.86 \\
\bottomrule[1pt]

\end{tabular}
\end{center}
\caption{Top-1 accuracy under the linear evaluation on ImageNet with the ResNet-50 backbone. The table compares the methods over 200 epochs of pretraining.
\# Parameters: the parameter counts of negative samples, as to BYOL and SimSiam, it means the parameter counts of the predictor. }
\label{tab:memory_size}
\end{table*}

\subsection{Results on Imagenet1K}

First, we perform a linear evaluation by fine-tuning a fully connected classifier for $100$ epochs on top of the frozen 2048-D feature vector from the ResNet-50 backbone. Table~\ref{tab:imagenet200} reports the results on Imagenet1K after $200$ epochs of pre-training. With different batch sizes of $256$, $1024$ and $4,096$ for network pre-training, CaCo even outperforms the state-of-the-art SWAV, BYOL and NNCLR with a much larger batch size of $4096$ when we used small batch-size. It is well known that the performances of these SOTA models improve with larger batch sizes \cite{caron2020unsupervised,grill2020bootstrap,dwibedi2021little}. However, larger batch sizes require multiple GPU servers with at least $32$ Nvidia V100 GPU cards to accommodate $4096$ images. This makes the network pre-training unaffordable and inconvenient to set up across multiple servers for most of research teams.  The proposed CaCo not only outperforms SOTA methods with much smaller batch sizes, but also allows affordable network pre-training on a single GPU server. Our training time is on par with AdCo~\cite{hu2021adco}, MoCo~\cite{he2020momentum} as shown in Table~\ref{tab:time_cost}

For single-crop settings, we also report the results with more than $200$ pre-training epochs in Table~\ref{tab:imagenet800}. Here we run the CaCo with a batch size of $1,024$ and $4,096$ images to compare with the other methods. Here $1,024$ is the largest size that can fit the GPU memory of a single $8\times V100$ server. It still outperforms the SOTA models with various batch sizes and epochs. % using even more epochs (e.g., up to $1,000$ epochs) and/or larger batch sizes.
This makes the CaCo affordable and competitive for network pre-training.
%In future, we will run and test the CaCo with larger batch sizes across multiple GPU servers  when they become available and affordable.
It is worth noting that here we do not apply any tricks such as multi-crop augmentations \cite{caron2020unsupervised,xu2021k} and stronger augmentations \cite{wang2021contrastive} that have shown effective in improving the accuracy in unsupervised learning literature. This shows the CaCo is an elegant method easier to implement without relying on those tricks. 

To compare with the methods using multi-crop augmentations \cite{caron2020unsupervised}, we have conducted the experiments following the same protocol used in SwAV~\cite{caron2020unsupervised}. The results are shown in Table.~\ref{tab:imagenetmulti}. We outperformed the other methods in this setting even with a smaller batch size of $2,048$ based on the same number of four 8-Nvidia Tesla V100 GPU servers. 
We believe the performance $75.7\%$ can be further boosted if we can increase the batch size to $4,096$ as used in other methods. However, this requires doubling computing resources, which makes it unaffordable for many research groups.

\subsection{Transfer Learning Results}

We also perform transfer learning tasks on various datasets\footnote{The link to Birdsnap dataset is broken or refers to a dataset with missing data, making it difficult to make a direct comparison.}. The ResNet50 backbone was pretrained on Imagenet1K over $800$ epochs with a batch size of $4096$. By following the linear evaluation protocol in \cite{grill2020bootstrap}, a fully connected layer of classifier is trained upon the frozen backbone.

Table~\ref{tab:transfer} reports the results under the linear evaluation compared with the other models. The results show the CaCo performs the best on $8$ out of these $11$ datasets among the compared self-supervised methods. 
%It also outperforms the fully supervised ResNet50 features pre-trained with Imagenet labels on xx out of the 11 datasets.
This demonstrates the CaCo representation pre-trained on the Imagenet1K dataset can well generalize to downstream tasks on a variety of datasets.

\subsection{Ablation Studies}
In this section, we present a thorough analysis of CaCo. Here we focused on two most important factors of CaCo: the size of the memory bank (the sum of the positives and negatives) and the contribution of positives and negatives to the performance. 

\textbf{Size of the memory banks.} We set up the experiments with batch-size 4096 with 200 epoch pre-training, which indicates the number of positive is always 4096. We only change the memory bank size across different experiments. As shown in Table~\ref{tab:memory_size}, we changed the memory size from 4M to 12M and observed the clear improvement of performance from $71.7\%$ to $72.1\%$. That's to say, CaCo can be benefited by the size increase of memory bank. This is reasonable since bigger memory bank size has two benefits: 1) more accurate positives by checking the most probable positive from a bigger memory bank; 2) the task become more challenge since encoder needs to select the correct positive out of a bigger pool including more negative pairs.

\textbf{Contribution of positives and negatives.} We conducted an ablation study to compare the contribution of positives and negatives. Here all experiments are based on a batch size of 1024 with 200 epoch pre-training. Here None setting is without memory bank, while both positive and negatives directly coming from the concurrent mini-batch. Here Positive setting is similar to that NNCLR~\cite{dwibedi2021little}, where we keep a queue to save the embeddings of previous batches and we select the nearest neighbor of query in memory bank as positive while negatives are from the concurrent mini-batch. Here Negative setting is similar to AdCo~\cite{hu2021adco}, where the memory bank is adversarially trained and serves as the negatives, and the positives are from the current mini-batch. Here P+N setting is our CaCo setting, where both positives and negatives are from the memory bank and memory bank is collaboratively and adversarially trained by the contrastive loss in an end-to-end fashion. We keep all other settings to the same to have a fair comparison. The results are shown in Table~\ref{tab:positive_nega}. It's clear that CaCo clearly improved compared to using positive or negative alone. 

\begin{figure*}[tb!]
%\captionsetup[subfigure]{singlelinecheck=false}
    \centering
    \begin{minipage}[c]{0.32\linewidth}
    \begin{center}
        \subfloat[CIFAR10]{\includegraphics[width=\textwidth]{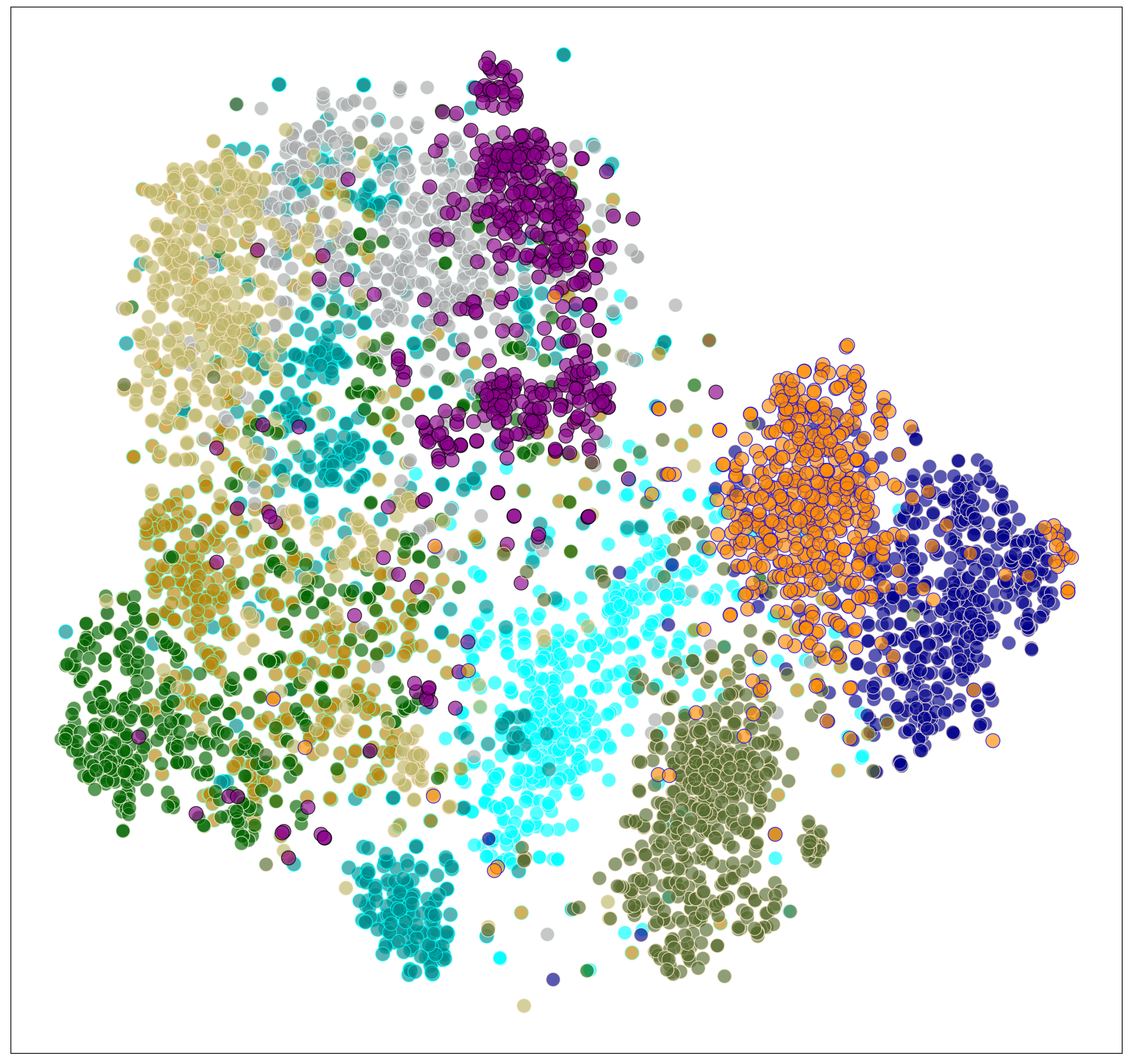} }
    \end{center}
    \end{minipage}
    \begin{minipage}[c]{0.32\linewidth}
    \begin{center}
        \subfloat[PETS]{\includegraphics[width=\textwidth]{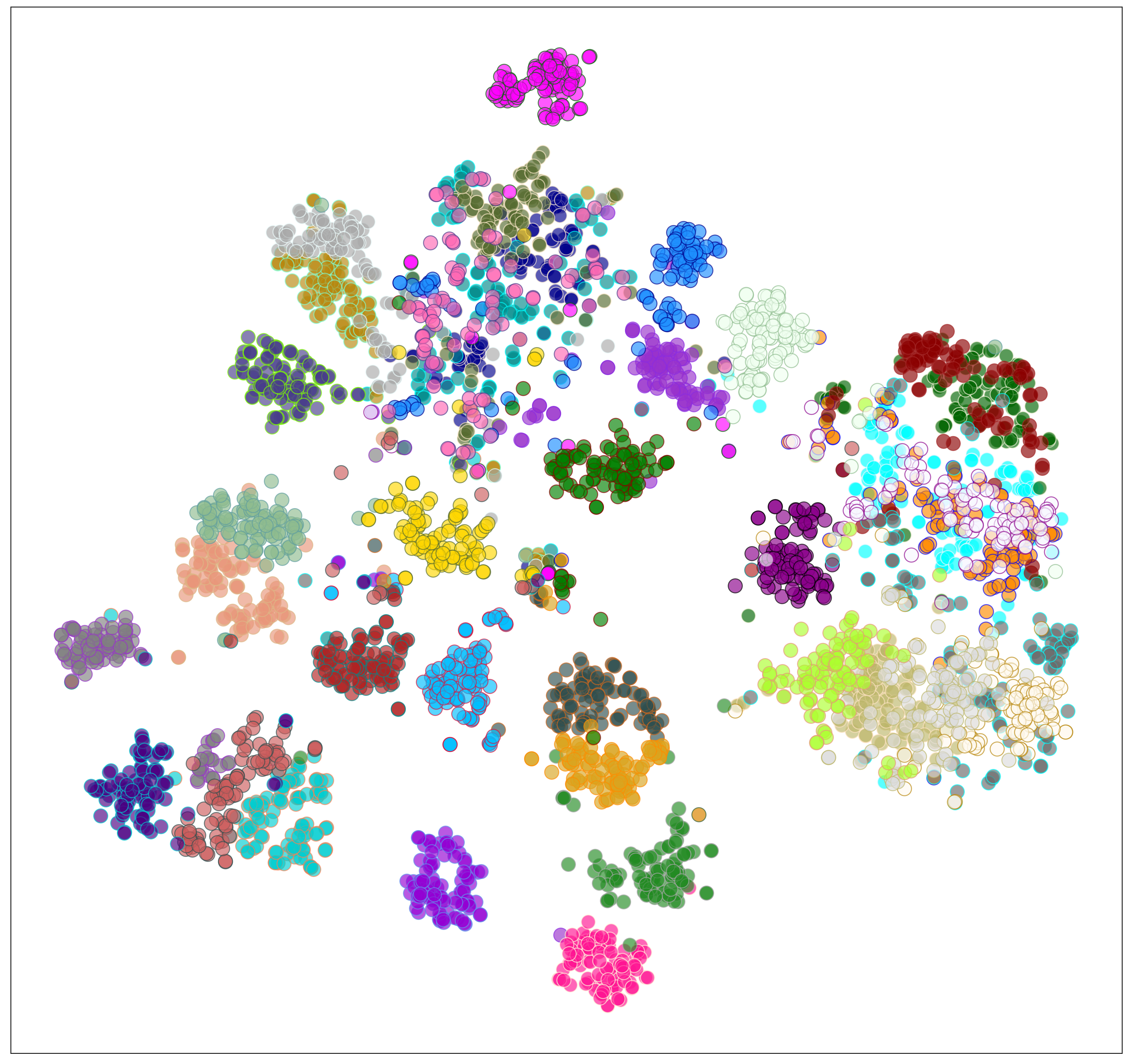}}
    \end{center}
    \end{minipage}
    \begin{minipage}[c]{0.32\linewidth}
    \begin{center}
        \subfloat[VOC2017]{\includegraphics[width=\textwidth]{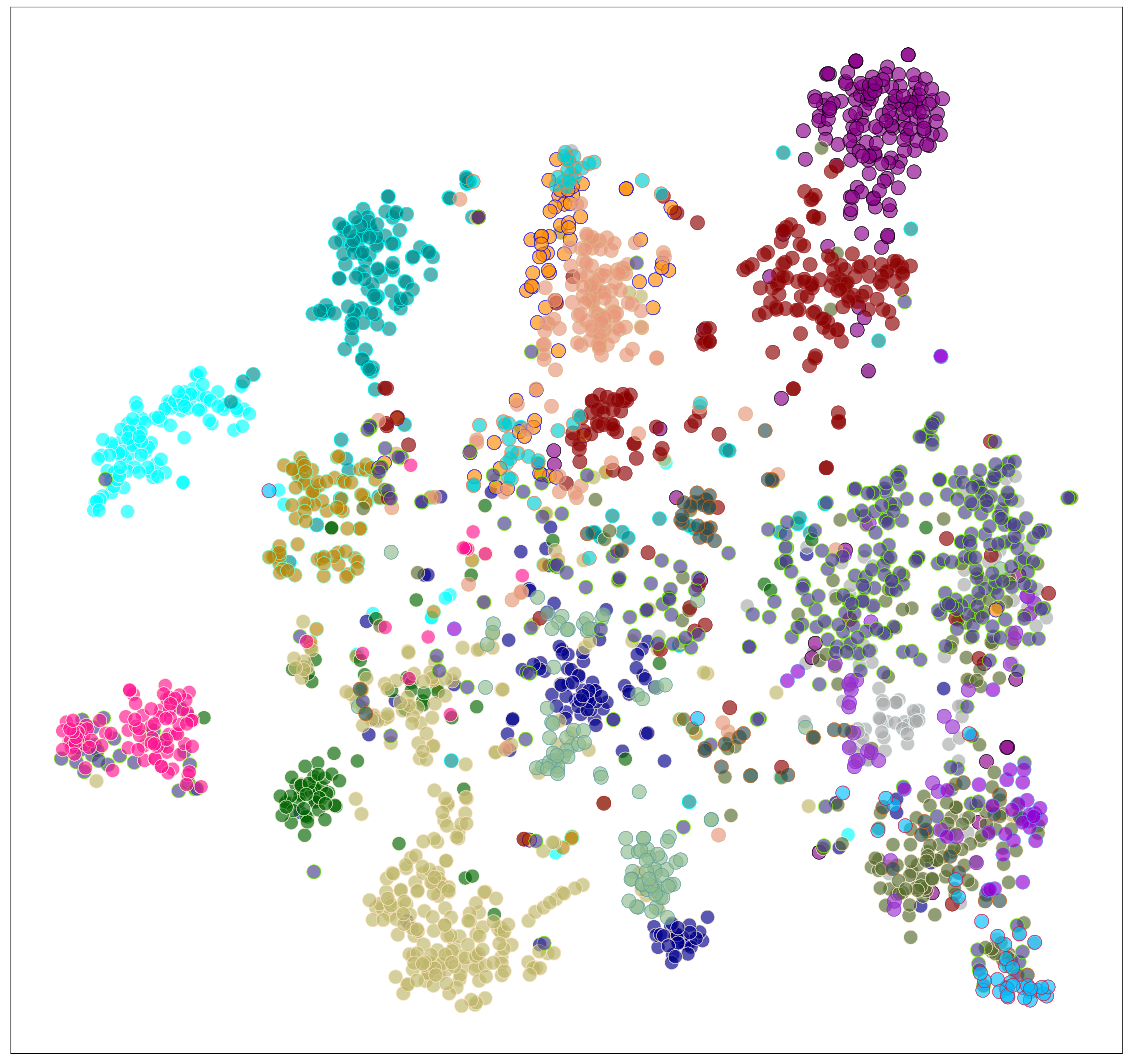}}
    \end{center}
    \end{minipage}
    \caption{t-SNE visualization of learned representations on CIFAR10, PETS and VOC2017 datasets, where different colors represent different classes on these datasets. The encoder is pre-trained by CaCo over 800 epochs on Imagenet1K. }
    %By comparison, MoCo trains neither negatives nor positives, while AdCo only trains adversarial negatives.
    \label{fig:t_sne}
   
\end{figure*}

%##################################################################################################
\begin{table*}
\begin{center}
\begin{tabular}{ccccc}
\toprule[1pt]
Setting &  Batch Size & \#Positive Pool &\#Negative Pool & Top-1 \\ 
\midrule[1pt]
None&1024&1024&1024&70.2\\
Positive&1024&65536&1024&70.7\\
Negative&1024&1024&65536&70.6\\
\midrule
Positive+Negative&1024&65536&65536&71.3\\
\bottomrule[1pt]

\end{tabular}
\end{center}
\caption{Top-1 accuracy under the linear evaluation on ImageNet with the ResNet-50 backbone. The table compares different settings over 200 epochs of pretraining. Here \textit{None} setting denotes both positive and negative are from current mini-batch; \textit{Positive} setting denotes the positive embedding is from memory bank and the negative embedding is from the current mini-batch; \textit{Negative} setting suggests the positive embedding is from current mini-batch and the negative embedding is from the memory bank; \textit{P+N} is our CaCo setting, where both positive and negative are from the memory bank. Here the \#Positive Pool and \#Negative Pool suggests the total number of embeddings that we selected from to serve as positive and negatives, respectively. }
\label{tab:positive_nega}
\end{table*}

%##################################################################################################
%\subsection{Results on PASCAL VOC and COCO}
\subsection{Result Analysis}

\subsubsection{Quality of Learned Positives and Negatives}
The learned positives and negatives in memory bank are ``virtual" samples that do not necessarily correspond to the features of some known images. To understand a virtual sample, one can find its most similar image from Imagenet in the feature space and use it as the surrogate.

We randomly choose some images as queries, and show the surrogates of the most probable positive samples (cf. Eq.~(\ref{eq:max_prob})) as well as the negative samples from the memory bank in Figure~\ref{fig:query_ex}. We can see that the positives (negatives) belong to the same (different) class of query images. This demonstrates that the queries successfully learn the true positives and true negatives from the memory bank.

Moreover, given a query, the maximum probability corresponding to the most probable positive in (\ref{eq:max_prob}) measures the likelihood of obtaining a true positive by the model. We plot the curve of the mean maximum positive probability over epochs in Figure~\ref{fig:prob_acc}, along with the nearest neighbor (NN) accuracy based on $20\%$ Imagenet1K training set. We plot the NN accuracy to measure the performance of the pre-trained representation as it can be computed in a few seconds.

We can see that at the beginning, the probability is close to zero as randomly choosing a positive sample from the memory bank of $65,536$ samples is as low as $1.5\times 10^{-5}$. After $200$ epochs of pretraining, the probability increase significantly by $10^3$ times to above $1.2\times 10^{-2}$. The increase in the mean maximum positive probability is consistent with the increase in the NN accuracy. It suggests that high-quality positives and negatives are eventually learned, leading to better representations learned over epochs, and this partly explains the success of the CaCo method.

%These findings show that high-quality positives and negatives are eventually learned, which partly explains the success of the CaCo method.

\subsubsection{Visualization of Representations}

In Figure~\ref{fig:t_sne}, we also use t-SNE to visualize the feature embeddings on CIFAR10, PETS and VOC2017 datasets. The encoder is learned by CaCo over 800 epochs on Imagenet1k dataset without labels. It shows that the feature embeddings obtained via t-SNE exhibit distributional patterns well aligned with the classes of various colors on these datasets.  This suggests  the learned unsupervised features on one dataset well generalize to downstream datasets without access to their labels.

% learned by the learned encoder of MoCo v2, AdCo and CaCo on CIFAR-10 dataset in Figure~\ref{}. These encoders are pre-trained on Imagenet1k dataset without labels in a self-supervised fashion over $800$ epochs. The 2D CaCo features obtained via t-SNE exhibit clustering patterns best aligned with the CIFAR-10 classes of various colors.  This visualizes how the unsupervised features well generalize to a downstream dataset without access to its labels.

\section{Conclusions}\label{sec:concl}
This paper presents Cooperative-adversarial Contrastive (CaCo) learning for network pre-training. It directly learns both positives and negatives over a shared memory bank as cooperators and adversaries to an encoder by minimizing and maximizing the underlying contrastive loss, respectively.
%where both positives and negatives share a memory bank, with the most probable and stable positives directly learned over the bank.
This results in an end-to-end training of positives and negatives iteratively updated towards query anchors along the directions tangent to the unit hypersphere. Jointly trained with the encoder, unsupervised features are learned to distinguish the learned adversarial negatives from the most probable cooperative positive given each query anchor. Experiments demonstrate that the CaCo outperforms the SOTA self-supervised methods on multiple downstream tasks without relying on multi-crop augmentations.

\appendices


\ifCLASSOPTIONcompsoc
  \section*{Acknowledgments}
   W. Xiao and Y. Huang implemented the idea and performed the experiments as remote research interns in the Seattle Research Center. G.-J. Qi was the corresponding author (e-mail: guojunq@gmail.com), who conceived and formulated the idea as well as wrote the paper.
\else
  \section*{Acknowledgment}
\fi

\ifCLASSOPTIONcaptionsoff
  \newpage
\fi



\bibliographystyle{IEEEtran}
\bibliography{reference}

\end{document}